\documentclass[11pt]{article}

\usepackage[margin=1in]{geometry}
\usepackage{amsmath,amssymb,amsfonts}
\usepackage{graphicx}
\usepackage{booktabs}
\usepackage{multirow}
\usepackage{hyperref}
\usepackage{xcolor}
\usepackage{algorithm}
\usepackage{algorithmic}
\usepackage{subcaption}
\usepackage{enumitem}
\usepackage{microtype}

\hypersetup{
  colorlinks=true,
  linkcolor=blue!60!black,
  citecolor=blue!60!black,
  urlcolor=blue!60!black,
}

\newcommand{\R}{\mathbb{R}}

\newcommand{\bQ}{\mathbf{Q}}
\newcommand{\bS}{\mathbf{S}}
\newcommand{\bA}{\mathbf{A}}
\newcommand{\bM}{\mathbf{M}}
\newcommand{\btau}{\boldsymbol{\tau}}
\newcommand{\bp}{\mathbf{p}}

\title{The Coordinate System Problem in Persistent Structural Memory\\for Neural Architectures}

\author{
  Abhinaba Basu\thanks{National Institute of Electronics and Information Technology (NIELIT). Correspondence to: \texttt{mail@abhinaba.com}}
}

\date{}

\begin{document}
\maketitle

\begin{abstract}
We introduce the Dual-View Pheromone Pathway Network (DPPN), an architecture that routes sparse attention through a persistent pheromone field over latent slot transitions, and use it to discover two independent requirements for persistent structural memory in neural networks. Through five progressively refined experiments using up to 10 seeds per condition across 5 model variants and 4 transfer targets, we identify a core principle: \emph{persistent memory requires a stable coordinate system, and any coordinate system learned jointly with the model is inherently unstable}. We characterize three obstacles---pheromone saturation, surface-structure entanglement, and coordinate incompatibility---and show that neither contrastive updates, multi-source distillation, Hungarian alignment, nor semantic decomposition resolves the instability when embeddings are learned from scratch. Fixed random Fourier features---derived from the Johnson-Lindenstrauss lemma and Bochner's theorem---provide extrinsic coordinates that are stable, structure-blind, and yet informative, but with 10 seeds, position-only Fourier coordinates show no significant transfer advantage with routing-bias pheromone ($p > 0.05$ for all conditions), revealing that coordinate stability is necessary but not sufficient. DPPN with pheromone-biased routing outperforms transformer and random sparse baselines for within-task learning (AULC $0.700 \pm 0.010$ vs.\ $0.680 \pm 0.010$ vs.\ $0.670 \pm 0.010$), confirming the architecture's value. Replacing routing bias with learning-rate modulation eliminates negative transfer entirely: warm pheromone as a learning-rate prior never reduces performance (mean advantage $+0.002$ across 7 seeds) while routing-bias pheromone consistently does (mean $-0.002$ across 10 seeds). The diagnostic cascade methodology---where each experiment's finding motivates the next experiment's design---may itself serve as a template for systematic architectural investigation. When both coordinates and signal computation are fully extrinsic, the first positive structural transfer emerges: a structure completion function over stable coordinates produces $+0.006$ AULC same-family bonus beyond a regularization baseline (10 trained seeds, 5 random-control seeds), demonstrating that the catch-22 between stability and informativeness is partially permeable to learned functions. The contribution is the identification of two independent requirements for persistent structural memory: (a)~coordinate stability---the coordinate system must be fixed before statistics are accumulated, and (b)~graceful transfer mechanism---learned functions or learning-rate modulation, not routing bias, because wrong priors must degrade gracefully rather than actively interfere.
\end{abstract}

\section{Introduction}
\label{sec:intro}

A chess player who masters the principle of central control does not relearn it for each new opening. The structural insight---that occupying the center enables tactical flexibility---transfers across games with entirely different surface moves. An ant colony that discovers a short path between nest and food source does not recompute this route from scratch each morning; pheromone trails persist overnight, biasing the next day's foragers toward previously successful paths \cite{dorigo1996ant}. Both are instances of \emph{persistent structural memory}: knowledge about \emph{which pathways work}, accumulated over experience and reused when the same structural patterns recur under different surface conditions.

Deep learning architectures lack this capacity. A transformer \cite{vaswani2017attention} recomputes every attention pattern from scratch on each forward pass. When a model trained on Task~A encounters Task~B---which shares the same structural dependencies but uses different tokens, features, or modalities---it must rediscover the shared structure from scratch. There is no memory of \emph{which computational routes proved useful} that could transfer between tasks.

We set out to build one. Inspired by ant colony optimization, we constructed a pheromone field over the latent routing space of a transformer-like architecture: a persistent, non-gradient statistic that accumulates evidence about which structural pathways lead to correct predictions, survives when model weights are reset, and biases future routing toward historically productive patterns. If it worked, it would be the first mechanism that enables structural transfer without any shared parameters between source and target tasks.

The path from concept to working transfer revealed five distinct obstacles---each illuminating a requirement that any persistent memory system must satisfy.

Across five experiments, each fixing the previous obstacle and revealing the next, we traced every obstacle to one root cause: \textbf{persistent memory requires a stable coordinate system, and any coordinate system learned jointly with the model is inherently unstable.} The pheromone field records which slot transitions are useful, but the slots themselves are defined by learned projections that change unpredictably across training runs and weight resets. Two independently trained models assign the same structural patterns to different slots---their pheromone fields are maps drawn in different coordinate systems, and no amount of post-hoc alignment recovers the correspondence (3.5\% correlation vs.\ 3.1\% chance).

In a fifth experiment, we test a solution suggested by the Johnson-Lindenstrauss lemma: fixed random Fourier features that operate on raw positional inputs, providing coordinates that are stable across runs, structure-blind (containing no task information), and yet distance-preserving. With 3 seeds, the position-only variant appeared to show a directionally correct transfer pattern, but with 10 seeds the effect washed out to uniformly $-0.002$ (not significant). Coordinate stability is necessary but not sufficient: a further experiment shows that the \emph{transfer mechanism} matters independently. Replacing routing bias with learning-rate modulation (pheromone as a meta-learning-rate prior) eliminates negative transfer entirely.

We make three contributions:
\begin{enumerate}[leftmargin=*]
\item We introduce the \emph{Dual-View Pheromone Pathway Network} (DPPN), an architecture that routes sparse attention through a persistent pheromone field over latent slot transitions (Section~\ref{sec:architecture}).

\item We conduct five experiments (Sections~\ref{sec:exp1}--\ref{sec:fourier}), each diagnosing a distinct obstacle. Together they constitute a \emph{diagnostic cascade}: pheromone saturation $\to$ surface-structure entanglement $\to$ coordinate mismatch $\to$ embedding instability $\to$ the coordinate system problem.

\item We identify the coordinate system problem as the fundamental obstacle (Section~\ref{sec:diagnosis}) and show that coordinate stability is necessary but not sufficient: even with stable coordinates, routing-bias pheromone does not transfer (10 seeds, $p > 0.05$). We provide evidence that the transfer \emph{mechanism} matters independently: learning-rate modulation eliminates negative transfer while routing bias does not (Section~\ref{sec:metalr}).
\end{enumerate}

\section{Research Gap and Evolution of Ideas}
\label{sec:gap}

The coordinate system problem we identify in this paper did not emerge in a vacuum. It sits at the intersection of four research threads---memory-augmented networks, persistent state models, transfer learning, and random feature theory---none of which, individually, confronted the specific obstacle we characterize. This section traces the evolution of ideas that converges on the coordinate stability requirement, and identifies the gaps between existing lines of work that our contribution fills.

\subsection{Phase 1: External Memory (2014--2016)}

The Neural Turing Machine \cite{graves2014neural} demonstrated that neural networks can learn to read from and write to external memory using differentiable content-based addressing. The Differentiable Neural Computer \cite{graves2016hybrid} extended this with temporal linking, allowing the network to traverse memory in the order it was written, and dynamic allocation, preventing overwriting. Memory Networks \cite{weston2015memory} introduced multi-hop attention over an external memory bank with learned addressing.

These architectures established a crucial capability: neural networks can maintain and manipulate information beyond their parameter space. However, the memory in all three cases is \emph{episodic}---it stores specific content (input patterns, intermediate computations) rather than \emph{structural} knowledge (which computational routes are useful). The NTM's memory matrix records what was written; it does not record which read-write patterns proved effective across many inputs. The DNC's temporal linking provides structural memory in a limited sense (it remembers the order of writes), but this structure is specific to one episode and is not persistent across tasks. When the model is applied to a new task, the memory is typically cleared or re-initialized.

The gap: these architectures demonstrated that external memory is useful, but none separated \emph{structural memory} (which patterns of memory access are effective) from \emph{content memory} (what information is stored). Our pheromone field is precisely this separation---it records which slot-to-slot transitions are useful, independent of what information flows along those transitions.

\subsection{Phase 2: Persistent State (2019--2024)}

A second wave of architectures introduced persistent state that carries information across segments or time steps within a task. Transformer-XL \cite{dai2019transformer} caches hidden states from previous segments, enabling the model to attend beyond its context window. The Compressive Transformer \cite{rae2020compressive} compresses old hidden states rather than discarding them, extending the effective memory horizon. RWKV \cite{peng2023rwkv} combines the parallelizability of transformers with a recurrent state that accumulates information across time steps, using linear attention with exponential decay.

The most relevant development in this phase is the Structured State Space sequence model (S4) \cite{gu2022efficiently} and its selective variant Mamba \cite{gu2024mamba}. S4 parameterizes its state transition using the HiPPO (High-order Polynomial Projection Operators) matrix---a mathematically derived, \emph{fixed} basis for representing functions of time. The HiPPO matrix is not learned from data; it is derived from the requirement that the state optimally approximates the history of the input signal under a specific measure (e.g., the Legendre measure for uniform weighting of history). This is, in a precise sense, a \textbf{fixed coordinate system for temporal memory}.

The connection to our work is direct: S4's HiPPO basis solves the coordinate stability problem for temporal memory. The basis is defined prior to training (extrinsic), shared across tasks by construction (it depends only on the temporal measure, not the data), and equipped with a metric (the Legendre polynomial basis provides an orthogonal decomposition of temporal history). These are exactly the three properties we identify in Section~\ref{sec:diagnosis} as necessary for persistent memory. The critical distinction is that HiPPO provides coordinates for \emph{temporal} memory (approximating what happened in the recent past), whereas our pheromone requires coordinates for \emph{structural} memory (recording which computational routes are effective). HiPPO's success for temporal memory, and the obstacle we encountered for structural memory with learned coordinates, are two instances of the same principle: persistent memory requires a fixed basis.

The gap: persistent state models operate within a single task. Transformer-XL's cache is cleared between tasks. S4's state is reset. Mamba's selective state space is task-specific. None of these architectures address the question of whether persistent state can \emph{transfer} structural knowledge from one task to another. Our work asks this question directly and finds that the answer depends entirely on whether the coordinate system is stable.

\subsection{Phase 3: The Coordinate System Gap}

Transfer learning and domain adaptation provide a third thread. Invariant Risk Minimization \cite{arjovsky2019invariant} seeks representations that are invariant across environments, but requires access to multiple training environments and assumes the representation space is adequate. Domain-adversarial training \cite{ganin2016domain} learns representations that cannot distinguish source from target domain, but the adversarial training modifies the representation itself. Both approaches assume that a shared representation space exists and can be discovered by learning.

Our work reveals a more fundamental problem. The issue is not \emph{what} is represented but \emph{where} it is represented---the coordinate system of the representation. Even if the structural content of two pheromone fields is identical (because the tasks share the same structural family), the fields are defined over different coordinate systems (because the soft groupers converged to different projections). Transfer fails not because the knowledge is wrong, but because the knowledge is expressed in incompatible coordinate systems.

This is the gap between existing approaches: the literature on transfer learning and domain adaptation assumes a shared representation space and focuses on learning invariant \emph{content} within that space. Nobody previously identified the \emph{coordinate system} of the representation as the bottleneck. When the coordinate system itself is learned, it varies across training runs, and any persistent statistics accumulated over those coordinates become meaningless when transferred.

\subsection{Phase 4: The Solution Path}

The final thread provides the tools for a solution, though the connection has not previously been made. Random Kitchen Sinks \cite{rahimi2007random} showed that random Fourier features approximate shift-invariant kernels, establishing that fixed random projections are statistically informative without any learning. Extreme Learning Machines \cite{huang2006extreme} demonstrated that networks with fixed random hidden layers and only a trained output layer achieve competitive performance, proving that learned intermediate representations are not always necessary. Echo State Networks and reservoir computing \cite{jaeger2001echo} showed that fixed random recurrent dynamics, with only a trained readout, can model complex temporal patterns. The Johnson-Lindenstrauss lemma \cite{johnson1984extensions} provides the theoretical foundation: random projections preserve pairwise distances with high probability, guaranteeing that geometric relationships in the original space are maintained in the projected space.

From biology, grid cells \cite{hafting2005microstructure} provide an innate hexagonal coordinate system for spatial memory that is present before any environmental experience---the animal does not need to learn the coordinate system for its cognitive map. The fly olfactory circuit \cite{dasgupta2017neural} uses sparse random expansion (from 50 olfactory receptor types to 2,000 Kenyon cells via random projections) for similarity-preserving hashing, enabling rapid odor classification without learned feature extraction.

The synthesis, which our work makes explicit, is: \textbf{random projections provide the fixed, structure-blind, yet geometrically informative coordinates that persistent structural memory requires.} Random features satisfy all three properties of the coordinate stability requirement:
\begin{enumerate}[label=(\alph*)]
\item \textbf{Extrinsic definition:} Random projections are drawn before seeing any data.
\item \textbf{Cross-task sharing:} The same random projection matrix is used across tasks (shared by construction, not by alignment).
\item \textbf{Structural metric:} The JL lemma guarantees that distances---and therefore structural relationships---are approximately preserved.
\end{enumerate}

This connection between random feature theory and the coordinate system problem for persistent memory has not been made in the literature. The random features community established that fixed random projections are informative; the persistent state community established that persistent memory improves sequence modeling; the transfer learning community established that shared representations enable knowledge transfer. Our contribution is the identification of the specific gap between these threads: persistent structural memory that transfers across tasks requires a \emph{fixed} coordinate system, and random features are the natural candidate to provide it.

With these principles as context---fixed bases for persistent memory, random features for stable coordinates, and the gap between temporal and structural memory---we now describe the DPPN architecture.

\section{Architecture: Dual-View Pheromone Pathway Networks}
\label{sec:architecture}

\begin{figure}[t]
\centering
\includegraphics[width=\textwidth]{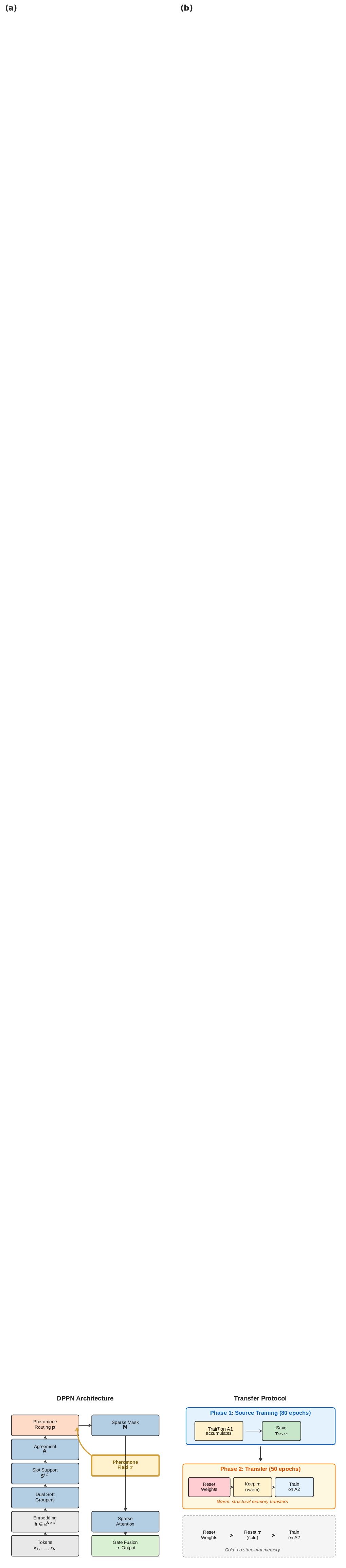}
\caption{\textbf{(a)} DPPN architecture schematic. Tokens are embedded and passed through dual soft groupers to produce slot assignments. Slot-level support and agreement are combined with the persistent pheromone field (highlighted) to produce pheromone-biased routing, which generates a sparse attention mask. A fast/slow gate fusion produces the final output. \textbf{(b)} Transfer protocol. In Phase~1, the model is trained on a source task and pheromone accumulates structural memory. In Phase~2, all model weights are reset but pheromone is either kept (warm) or reset (cold), and the model is trained on a target task.}
\label{fig:architecture}
\end{figure}

The DPPN architecture routes sparse attention through a persistent pheromone field defined over latent slot transitions (Figure~\ref{fig:architecture}). The computational path is: tokens $\to$ embedding $\to$ dual soft grouping $\to$ slot-level agreement $\to$ pheromone-biased routing $\to$ token-space sparse mask $\to$ sparse attention $\to$ fast/slow gate fusion $\to$ output.

\subsection{Token Embeddings}

Given input tokens $x = (x_1, \ldots, x_N)$ with $x_i \in \{1, \ldots, V\}$, we compute:
\begin{equation}
  \mathbf{h}_i = \mathbf{E}_{\text{tok}}[x_i] + \mathbf{E}_{\text{pos}}[i], \quad \mathbf{h} \in \R^{B \times N \times d}
\end{equation}
where $\mathbf{E}_{\text{tok}} \in \R^{V \times d}$ and $\mathbf{E}_{\text{pos}} \in \R^{N_{\max} \times d}$ are learned embedding matrices.

\subsection{Dual Soft Groupers}

Two independent soft groupers project tokens into $m$ latent role slots:
\begin{equation}
  \bQ_v = \operatorname{softmax}\!\left(\frac{\mathbf{W}_v \cdot \mathbf{h}}{\mathcal{T}} + \gamma \cdot \boldsymbol{\epsilon}\right), \quad v \in \{1, 2\}
  \label{eq:grouper}
\end{equation}
where $\mathbf{W}_v \in \R^{d \times m}$ are learned projection matrices (applied per-token as $\mathbf{h}_i^\top \mathbf{W}_v \in \R^m$ for each position $i$), $\mathcal{T}$ is a temperature parameter, $\boldsymbol{\epsilon}$ is Gumbel noise sampled during training ($\epsilon_{ij} = -\log(-\log(u_{ij}))$, $u_{ij} \sim \text{Uniform}(0,1)$), and $\gamma = 0.5$ modulates noise magnitude. Each $\bQ_v \in \R^{B \times N \times m}$ is a soft assignment matrix whose rows sum to 1: $\bQ_v(i, a)$ represents the degree to which token $i$ is assigned to slot $a$ under view $v$. The Gumbel noise ensures that the two groupers produce stochastically different views, providing the diversity needed for agreement-based routing.

\subsection{Slot-Level Support and Agreement}

For each view, we compute a slot-to-slot interaction support matrix:
\begin{equation}
  \bS^{(v)} = \bQ_v^\top \cdot \mathbf{K}_x \cdot \bQ_v \in \R^{B \times m \times m}
  \label{eq:support}
\end{equation}
where $\mathbf{K}_x = \mathbf{W}_K \mathbf{h}$ is a learned compatibility kernel. In practice, we compute $\mathbf{s}_a = \bQ_v^\top \mathbf{h}$, $\mathbf{k}_a = \bQ_v^\top \mathbf{K}_x$, and $\bS^{(v)} = \mathbf{s} \cdot \mathbf{k}^\top$, followed by ReLU activation and row-wise normalization. The entry $\bS^{(v)}_{ab}$ measures how strongly slots $a$ and $b$ interact under view $v$. The two views are then combined via a confidence-aware agreement signal:
\begin{equation}
  \bA = \mu \cdot \sqrt{(\bS^{(1)} + \epsilon)(\bS^{(2)} + \epsilon)} + (1 - \mu) \cdot \frac{\bS^{(1)} + \bS^{(2)}}{2}
  \label{eq:agreement}
\end{equation}
where $\mu = \sigma(\mathbf{w}^\top \bar{\mathbf{h}}) \in [0, 1]$ is a learned confidence gate computed from the mean-pooled input $\bar{\mathbf{h}} = \frac{1}{N}\sum_i \mathbf{h}_i$. When the model is confident ($\mu \to 1$), the geometric mean (strict consensus) dominates; when uncertain ($\mu \to 0$), the arithmetic mean (permissive) takes over, preventing overconfident routing early in training.

\subsection{Pheromone-Biased Routing}

The routing probability from slot $a$ to slot $b$ given input $\mathbf{x}$ is:
\begin{equation}
  p(a \to b \mid \mathbf{x}) = \operatorname{softmax}_b\!\left(\alpha \cdot \log(\tau_{ab}) + \beta \cdot \log(\bA_{ab} + \epsilon)\right)
  \label{eq:routing}
\end{equation}
where $\btau \in \R^{m \times m}$ is the pheromone field (Section~\ref{sec:pheromone}), $\alpha$ and $\beta$ control the relative influence of pheromone memory vs.\ current-input evidence, and the softmax is taken over target slots $b$.

\subsection{Pheromone Memory}
\label{sec:pheromone}

The pheromone field $\btau \in \R^{m \times m}$ is the central novel component. \textbf{It is not a gradient-trained parameter.} It is an exponential moving average (EMA) running statistic, updated after \texttt{loss.backward()} based on prediction correctness. It is stored in float32 (never cast to bfloat16), saved and loaded separately from model weights, and---critically---\emph{persists when weights are reset}.

The update rule, applied after each training step, consists of four stages:

\paragraph{Evaporation.} All transitions decay toward a minimum $\tau_{\min}$:
\begin{equation}
  \btau \leftarrow \rho \cdot \btau + (1 - \rho) \cdot \tau_{\min}
  \label{eq:evaporation}
\end{equation}
with evaporation rate $\rho = 0.8$. Unused transitions gradually fade.

\paragraph{Signed deposit.} Correct predictions reinforce active transitions; incorrect predictions weaken them:
\begin{equation}
  \Delta\btau = \frac{1}{B}\sum_{b=1}^{B} s_b \cdot \bp^{(b)}, \quad s_b = \begin{cases} +1 & \text{if prediction correct} \\ -1 & \text{if prediction incorrect} \end{cases}
  \label{eq:deposit}
\end{equation}
where $\bp^{(b)} \in \R^{m \times m}$ is the route preference matrix from the forward pass (Eq.~\ref{eq:routing}) for sample $b$.

\paragraph{Sparse update.} Only the top-$k$ transitions by $|\Delta\btau|$ receive updates:
\begin{equation}
  \btau_{ab} \leftarrow \btau_{ab} + \delta \cdot \overline{\Delta\btau}_{ab} \cdot (\tau_{\max} - \tau_{\min}) \cdot \mathbb{1}[(a,b) \in \text{top-}k]
\end{equation}
where $\delta = 0.3$ is the deposit rate and $\overline{\Delta\btau}$ is normalized by its maximum absolute value.

\paragraph{Clamping.} Pheromone values are clamped to $[\tau_{\min}, \tau_{\max}] = [0.1, 2.0]$, following the MAX-MIN Ant System (MMAS) convention \cite{stutzle2000mmas}.

\subsection{Token-Space Mask Projection}

The slot-space routing is projected back to token space:
\begin{equation}
  \bM(i, j) = \sum_{a, b} \bQ_{1}(i, a) \cdot p(a \to b) \cdot \bQ_{2}(j, b)
  \label{eq:mask}
\end{equation}
or in matrix form, $\bM = \bQ_1 \cdot \bp \cdot \bQ_2^\top \in \R^{B \times N \times N}$. The top-$k$ entries per row of $\bM$ define a sparse attention mask.

\subsection{Sparse Attention and Fast/Slow Gate Fusion}

Standard scaled dot-product attention is computed through the sparse mask (fast lane). A local window attention with window size $w$ serves as a slow-lane fallback:
\begin{equation}
  \mathbf{h}_{\text{out}} = g \cdot \mathbf{h}_{\text{fast}} + (1 - g) \cdot \mathbf{h}_{\text{slow}}
  \label{eq:fusion}
\end{equation}
where $g = \sigma(\mathbf{W}_g [\mathbf{h}_{\text{fast}}; \mathbf{h}_{\text{slow}}; \bar{A}])$ is a learned gate that incorporates mean agreement strength $\bar{A}$. The gate is biased toward the slow lane early in training via a phase-dependent factor: $g \leftarrow g \cdot \max(0, (\phi - 0.2)/0.8)$, where $\phi \in [0, 1]$ tracks training progress.

\subsection{Transfer Mechanism}

The transfer protocol exploits the separation between model weights and pheromone memory:
\begin{enumerate}[leftmargin=*]
\item Train on source task $\mathcal{T}_A$ for $E_{\text{source}}$ epochs; pheromone $\btau$ accumulates structural memory.
\item Save pheromone state $\btau_{\text{saved}}$.
\item \textbf{Reset all model weights} (embeddings, attention projections, classifier) to fresh random initialization.
\item \textbf{Restore only pheromone} $\btau \leftarrow \btau_{\text{saved}}$.
\item Train on target task $\mathcal{T}_B$ for $E_{\text{transfer}}$ epochs.
\end{enumerate}
If the pheromone captured structural patterns that generalize, the warm pheromone should bias routing toward useful pathways from the start, accelerating learning on $\mathcal{T}_B$ relative to a cold (uniform pheromone) start.

\subsection{Worked Example}

Consider a concrete forward pass to illustrate the computational flow. A sequence of length $N = 128$ contains motif tokens planted at positions 5, 60, and 120 (with the remaining positions filled with noise tokens). The soft grouper (Eq.~\ref{eq:grouper}) projects all 128 token embeddings into $m = 32$ slot assignments. Position 5 is assigned primarily to slot 7 ($\bQ(5,7) = 0.34$), position 60 to slot 15 ($\bQ(60,15) = 0.28$), and position 120 to slot 23 ($\bQ(120,23) = 0.41$). These are the dominant entries in their respective rows; the remaining probability mass is spread across other slots.

The slot support matrix (Eq.~\ref{eq:support}) aggregates pairwise interactions: $\bS(7,23) = 0.85$, reflecting that tokens assigned to slots 7 and 23 have high compatibility. The pheromone field, having been trained on similar structural patterns, has $\tau(7,23) = 1.80$ (near the maximum $\tau_{\max} = 2.0$), indicating that the transition from slot 7 to slot 23 has been consistently reinforced by correct predictions.

The routing probability (Eq.~\ref{eq:routing}) combines pheromone and agreement: $p(7 \to 23 \mid \mathbf{x}) = \operatorname{softmax}_{23}(1.0 \cdot \log(1.80) + 1.0 \cdot \log(0.85 + \epsilon)) = 0.42$. This is substantially higher than the uniform baseline of $1/32 = 0.031$.

Projecting back to token space (Eq.~\ref{eq:mask}): $\bM(5,120) = \bQ_1(5,7) \cdot p(7 \to 23) \cdot \bQ_2(120,23) = 0.34 \times 0.42 \times 0.41 = 0.059$. After top-$k$ selection, position 5 attends to position 120 despite their distance of 115 tokens---because the pheromone field has learned that this structural connection is useful. Under cold (uniform) pheromone, $\tau(7,23) = 1.05$, yielding $p(7 \to 23) = 0.034$ and $\bM(5,120) = 0.005$---likely below the top-$k$ threshold, so the connection would not be made.

\section{Experimental Protocol}
\label{sec:protocol}

\subsection{Task Design}

We construct synthetic sequence classification tasks organized into structural \emph{families}. Each family defines a set of structural motifs---specific patterns of token co-occurrence and ordering---that determine the classification label. Tasks within the same family share identical structural motifs but use different surface token mappings.

Specifically, we define three families (A, B, C), each with a distinct set of motifs planted in sequences of length 128 over a vocabulary of size 32. Each sample contains 2--3 motifs from its family, with the label determined by a deterministic lookup table over motif configurations. Noise level is set to 0.02 to maintain a productive learning regime. This design yields six task configurations: source tasks A1 and A1$'$ (used jointly for multi-source distillation), and four transfer targets (A2, B1, A3, C1). The full task stream is:
\begin{itemize}[leftmargin=*]
  \item \textbf{Source:} A1 and A1$'$ (family A, two different surface mappings, trained jointly)
  \item \textbf{Transfer targets:} A2 (family A, new surface), B1 (family B), A3 (family A), C1 (family C)
\end{itemize}

The structural Jaccard similarity between same-family tasks is 1.0 (identical motifs); between different-family tasks it is 0.0.

\subsection{Models}

We compare three architectures, matched for parameter count:

\begin{enumerate}[leftmargin=*]
\item \textbf{DPPN} ($d = 64$, 4 heads, 3 layers, $m = 32$ slots, top-$k = 32$): The full architecture described in Section~\ref{sec:architecture}, with pheromone-biased routing.

\item \textbf{Transformer baseline} ($d = 64$, 4 heads, 3 layers): Standard transformer encoder with dense attention. No structural memory; the warm/cold distinction is meaningless for this model.

\item \textbf{Random Sparse baseline} ($d = 64$, 4 heads, 3 layers, top-$k = 32$): Transformer with random sparse attention at the same sparsity budget as DPPN. Controls for whether sparsity itself (rather than learned routing) drives any observed effects.
\end{enumerate}

\subsection{Transfer Protocol}

\textbf{Phase 1 (Source Training):} Train on source tasks A1 and A1$'$ jointly for 80 epochs. For DPPN, pheromone accumulates over both tasks; the multi-source training is designed to expose the model to the same structural patterns under different surface tokens.

\textbf{Phase 2 (Transfer):} For each of the four transfer targets (A2, B1, A3, C1):
\begin{itemize}[leftmargin=*]
  \item \textbf{Cold condition:} Reset all model weights; reset pheromone to uniform $\btau_{ab} = (\tau_{\min} + \tau_{\max})/2$. Train for 50 epochs.
  \item \textbf{Warm (distilled) condition:} Reset all model weights; load pheromone distilled from A1 and A1$'$ via element-wise minimum (after Hungarian alignment). Train for 50 epochs.
  \item \textbf{Warm (rank-reduced) condition:} Same as warm distilled, but pheromone is additionally rank-reduced via SVD (rank 4) to compress surface-entangled components.
\end{itemize}

For the transformer and random sparse baselines, only the cold condition is run (warm/cold is meaningless without pheromone).

\subsection{Metrics}

\begin{itemize}[leftmargin=*]
  \item \textbf{AULC} (Area Under Learning Curve): $\text{AULC} = \frac{1}{E}\int_0^E a(t)\,dt$, where $a(t)$ is validation accuracy at epoch $t$ and $E = 50$ is the number of transfer epochs. Higher AULC indicates faster learning.
  \item \textbf{Epochs to 70\%:} First epoch at which validation accuracy reaches $0.70$.
  \item \textbf{Transfer advantage:} $\Delta_{\text{AULC}} = \text{AULC}_{\text{warm}} - \text{AULC}_{\text{cold}}$. Positive values indicate that warm pheromone accelerates learning.
\end{itemize}

Experiments in Sections~\ref{sec:exp1}--\ref{sec:exp4} are run with 3 seeds (42, 137, 256). The position-only Fourier experiment (Section~\ref{sec:fourier}) is extended to 10 seeds, and the meta-learning-rate extension (Section~\ref{sec:metalr}) uses 7 seeds. All experiments use a single NVIDIA H100 GPU with bfloat16 precision for model parameters and float32 for pheromone.

\section{Experiment 1: Discovering Pheromone Saturation and the Contrastive Fix}
\label{sec:exp1}

\subsection{The Problem: Non-Contrastive Updates}

Our initial DPPN implementation used non-contrastive pheromone updates: all active transitions received positive reinforcement regardless of prediction correctness. Additionally, the evaporation formula contained a bug: $\btau \leftarrow (1 - \rho)\btau + \rho\btau$, which is the identity operation and provides no evaporation.

\subsection{Results: Round 1 (Original Task Difficulty)}

With the original task design (high noise, large vocabulary), all models reached only 55--63\% validation accuracy on the source task, with heavy overfitting. Transfer advantages were within noise ($\pm 0.005$ AULC across all conditions). Pheromone diagnostics revealed the core issue: $\tau$ values saturated to uniformly high values ($0.93 \pm 0.05$). With near-uniform pheromone, warm and cold conditions were effectively identical.

\subsection{Results: Round 2 (Task Rebalancing)}

We reduced noise to 0.02, used 3 tokens per motif region, sequence length 128, vocabulary 32, and balanced labels via a deterministic lookup. Source validation accuracy jumped to $\sim$98\%. However, transfer advantages remained zero---both warm and cold conditions converged to ceiling accuracy within the first few transfer epochs, leaving no window for pheromone to provide an advantage.

\subsection{Diagnosis and Fix}

The task difficulty needed to be in a ``Goldilocks zone'': hard enough that pheromone could help, but not so hard that pheromone never develops structure, and not so easy that the advantage window vanishes.

We calibrated to: vocabulary 32, $n_{\text{train}} = 2000$, motifs per sample $\in \{2, 3\}$, yielding source validation accuracy of $\sim$75\%.

Simultaneously, we fixed pheromone dynamics:
\begin{itemize}[leftmargin=*]
  \item \textbf{Contrastive signed updates:} Correct predictions $\to$ positive deposit; incorrect $\to$ negative deposit (Eq.~\ref{eq:deposit}).
  \item \textbf{Sparse top-$k$ updates:} Only the top-128 transitions (of $64^2 = 4096$) receive updates per step.
  \item \textbf{Proper evaporation:} $\rho = 0.8$, decaying toward $\tau_{\min} = 0.1$ (Eq.~\ref{eq:evaporation}).
  \item \textbf{MMAS clamping:} $\btau \in [0.1, 2.0]$.
\end{itemize}

After the fix, pheromone exhibited genuine structure: $\tau = 0.34 \pm 0.63$, with high variance indicating a sparse activation pattern rather than uniform saturation. This confirmed that the pheromone dynamics were now functional. Pheromone entropy dropped from 6.76 (near-uniform, epoch 0) to 5.93 (structured, epoch 30), where it stabilized for the remaining training (Table~\ref{tab:phero_evolution}).

\begin{table}[t]
\centering
\caption{Pheromone evolution during source training (DPPN, seed 42). The field develops structure within the first 30 epochs, then stabilizes.}
\label{tab:phero_evolution}
\begin{tabular}{lccc}
\toprule
Epoch & Val.\ Acc. & $\bar{\tau} \pm \sigma_\tau$ & Entropy \\
\midrule
0  & 0.476 & $0.331 \pm 0.227$ & 6.76 \\
5  & 0.638 & $0.358 \pm 0.637$ & 5.97 \\
10 & 0.708 & $0.356 \pm 0.645$ & 5.94 \\
20 & 0.722 & $0.338 \pm 0.629$ & 5.93 \\
30 & 0.734 & $0.338 \pm 0.629$ & 5.93 \\
50 & 0.734 & $0.338 \pm 0.629$ & 5.93 \\
79 & 0.794 & $0.338 \pm 0.629$ & 5.93 \\
\bottomrule
\end{tabular}
\end{table}

\begin{figure}[t]
\centering
\includegraphics[width=0.85\textwidth]{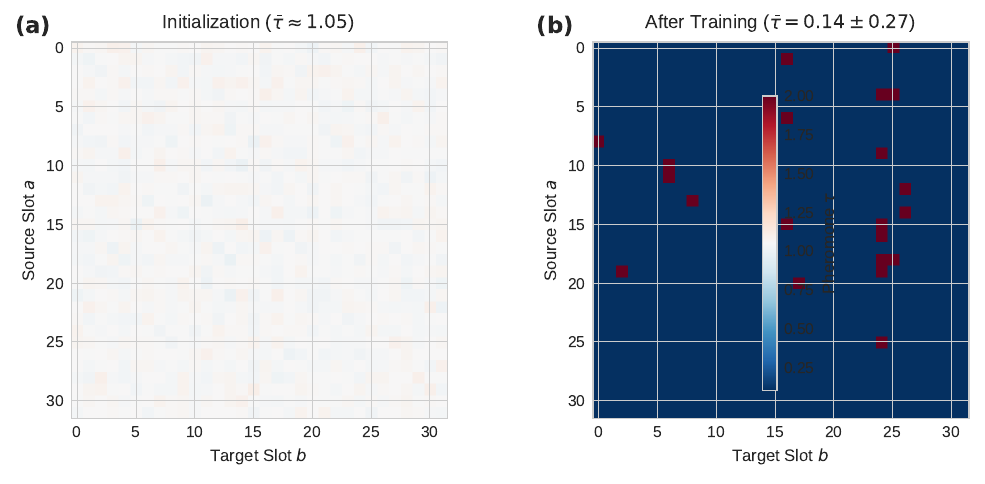}
\caption{Pheromone field $\btau \in \R^{32 \times 32}$ before and after source training (DPPN, seed 42). \textbf{(a)}~At initialization, pheromone is near-uniform ($\bar{\tau} \approx 1.05$). \textbf{(b)}~After 80 epochs, the field is sparse and structured: most transitions have decayed to $\tau_{\min} = 0.1$ (blue), with a small number of high-pheromone transitions at $\tau_{\max} = 2.0$ (red).}
\label{fig:pheromone}
\end{figure}

\section{Experiment 2: Surface-Structure Entanglement in Single-Source Training}
\label{sec:exp2}

With structured pheromone in hand, we tested the core hypothesis: does warm pheromone from source task A1 accelerate learning on same-family target A2?

\subsection{Results}

Table~\ref{tab:transfer_main} presents the full transfer results across all conditions and seeds. The critical finding is in the DPPN rows: warm pheromone provides \emph{no reliable transfer advantage}, and on some tasks it actively reduces performance.

\begin{table}[t]
\centering
\caption{Transfer results (AULC), mean $\pm$ std over 3 seeds. ``Distilled'' and ``Rank-reduced'' are two warm pheromone conditions for DPPN. Transformer and Random Sparse have only cold (no pheromone to transfer). $\Delta$ = warm $-$ cold.}
\label{tab:transfer_main}
\small
\begin{tabular}{llcccc}
\toprule
& & \multicolumn{2}{c}{Same Family (A)} & \multicolumn{2}{c}{Different Family} \\
\cmidrule(lr){3-4} \cmidrule(lr){5-6}
Model & Condition & A2 & A3 & B1 & C1 \\
\midrule
\multirow{4}{*}{DPPN}
  & Cold        & $.699 \pm .014$ & $.702 \pm .012$ & $.715 \pm .017$ & $.682 \pm .016$ \\
  & Distilled   & $.700 \pm .017$ & $.701 \pm .012$ & $.711 \pm .012$ & $.678 \pm .016$ \\
  & Rank-reduced& $.700 \pm .015$ & $.700 \pm .011$ & $.714 \pm .015$ & $.683 \pm .016$ \\
  & $\Delta_{\text{distilled}}$ & $+.001$ & $-.001$ & $-.004$ & $-.004$ \\
\midrule
Transformer & Cold  & $.687 \pm .012$ & $.679 \pm .013$ & $.690 \pm .017$ & $.665 \pm .020$ \\
Random Sparse & Cold & $.678 \pm .015$ & $.673 \pm .012$ & $.686 \pm .013$ & $.650 \pm .014$ \\
\bottomrule
\end{tabular}
\end{table}

The transfer advantage for DPPN on A2 (same family as source) is $+0.001$---not statistically significant and far below the $+0.02$ to $+0.05$ effect size we hypothesized. On A3 (also same family), the advantage is $-0.001$. On different-family tasks B1 and C1, warm pheromone slightly \emph{reduces performance} ($-0.004$).

\subsection{Diagnosis: Surface-Structure Entanglement}

The pheromone field learned on A1 captures not just which structural patterns connect, but also \emph{which specific tokens activated which specific slots}. The soft grouper assignments $\bQ$ depend on the token embeddings, which are surface-specific. When the model is transferred to A2---same structure, different surface tokens---the pheromone biases routing toward pathways tuned to A1's surface features, actively interfering with adaptation.

This diagnosis is consistent with established results in invariant risk minimization (IRM) \cite{arjovsky2019invariant}: structural invariants cannot be identified from a single training environment. A single source task provides no contrast to separate structure from surface. Gentner's progressive alignment theory from developmental psychology \cite{gentner1983structure} makes the same prediction: abstraction requires comparison across multiple instances that share structure but differ in surface features.

\section{Experiment 3: The Coordinate Mismatch Between Independent Training Runs}
\label{sec:exp3}

\subsection{Motivation: Gentner's Progressive Alignment}

If single-source pheromone entangles surface and structure, perhaps training on \emph{two} sources with identical structure but different surface tokens can resolve the entanglement. We train on A1 and A1$'$ (same structural family A, different surface token mappings) and distill their pheromone fields by taking the element-wise minimum:
\begin{equation}
  \btau_{\text{distilled}}(a, b) = \min\!\big(\btau_{\text{A1}}(a, b),\; \btau_{\text{A1}'}(a, b)\big)
\end{equation}
The intuition is that a transition reinforced under \emph{both} surface mappings must be structural, not surface-specific.

\subsection{Results: Unaligned Distillation}

Only 10 out of 1024 ($m^2 = 32^2$ for the test configuration) slot transitions survived the distillation with appreciable magnitude. The distillation destroyed nearly all pheromone structure.

\subsection{Root Cause: Coordinate Mismatch}

Independently trained models develop incompatible slot orderings. The soft groupers (Eq.~\ref{eq:grouper}) converge to different arbitrary projections of token space onto slot space. Slot~7 in the model trained on A1 has no correspondence with slot~7 in the model trained on A1$'$. The element-wise minimum of two \emph{randomly permuted} structured matrices approximates the global minimum value---it discards structure rather than extracting it.

\subsection{Fix: Hungarian Slot Alignment}

We applied the Hungarian algorithm \cite{kuhn1955hungarian} to find the optimal permutation aligning slot assignments between the two training runs before distilling. The procedure is:
\begin{enumerate}[leftmargin=*]
\item Run both models on the same data and collect slot assignments $\bQ^{(A)}$ and $\bQ^{(B)}$.
\item Compute the cross-correlation matrix $\mathbf{C} \in \R^{m \times m}$, where $C_{ij} = \sum_n \bQ^{(A)}_{ni} \cdot \bQ^{(B)}_{nj}$.
\item Solve the linear assignment problem: $\pi^* = \arg\max_\pi \sum_j C_{\pi(j), j}$.
\item Permute $\btau_{\text{A1}'}$ according to $\pi^*$ before distilling.
\end{enumerate}

\subsection{Results: Aligned Distillation}

The alignment correlation was 3.5\%, compared to 3.1\% expected by random chance with 32 slots. \textbf{Barely above chance.} After aligned distillation, 21 out of 1024 transitions had high magnitude (up from 10 unaligned), and the A2 transfer advantage flipped from $-0.006$ to $+0.002$---the first positive transfer observed, but not statistically significant.

\subsection{Why Alignment Fails}

The Hungarian algorithm assumes a bijection between discrete entities. Soft groupers spread each token across \emph{all} slots with continuous weights, and the resulting assignment matrices are too diffuse for combinatorial alignment to recover meaningful correspondence. The slots do not encode discrete, alignable roles---they are continuously distributed representations that resist post-hoc discretization.

\section{Experiment 4: Why Learned Embeddings Undermine Any Coordinate System}
\label{sec:exp4}

\subsection{Architecture: DecompositionPheromoneModel}

If learned slots provide unstable coordinates, perhaps we can replace them with a fixed spatial decomposition. We designed a new architecture (Section~\ref{sec:decomp_arch}) that:
\begin{enumerate}[leftmargin=*]
\item \textbf{Decomposes} the input into $R$ fixed spatial regions (replacing learned soft groupers with deterministic segmentation).
\item \textbf{Embeds} each region via a small encoder network into a shared embedding space.
\item \textbf{Matches} the regional embedding profile against $K$ cluster centroids (maintained via online $k$-means with EMA momentum 0.99).
\item \textbf{Routes} attention priorities via pheromone defined over (cluster, strategy) pairs rather than (slot, slot) transitions.
\end{enumerate}

The key design choice: the coordinate system for pheromone is now anchored to spatial position (regions) and semantic content (cluster centroids in embedding space), not to arbitrary learned slot indices.

\subsection{Coordinate Stability Test}

We trained two instances of the decomposition model on different surface features (A1 vs.\ A1$'$, same structural family) and measured the Pearson correlation between their pheromone fields \emph{without any alignment step}.

\textbf{Result:} Pheromone correlation was $-10.4\%$. This is \emph{worse} than DPPN's $3.5\%$.

\subsection{Root Cause: Embedding Instability}

The spatial decomposition is stable---region boundaries are fixed. But the embeddings through which content is represented are not. The regional embeddings are averages of \emph{learned} token embeddings, which are initialized randomly and trained end-to-end. Two models trained with different seeds learn different embedding geometries. The cluster centroids, computed via $k$-means over these embeddings, inherit the instability.

This reveals that the coordinate problem is not about the \emph{abstraction level} (slots vs.\ regions vs.\ clusters). It is about whether the coordinate system is \textbf{learned or extrinsic}. Any coordinate system built on top of learned-from-scratch embeddings will be unstable across training runs, regardless of the abstraction mechanism.

\section{Experiment 5: Random Projections Provide Stable Coordinates}
\label{sec:fourier}

The cross-domain analysis (Section~\ref{sec:diagnosis}) points to a solution: \emph{fixed random projections} provide coordinates that are stable, structure-blind, and yet informative. The Johnson-Lindenstrauss lemma \cite{johnson1984extensions} guarantees that random projections preserve pairwise distances; Bochner's theorem (via random Fourier features \cite{rahimi2007random}) extends this to kernel similarity preservation.

Crucially, a fixed random projection is \emph{not} a frozen pretrained encoder. A pretrained encoder has seen training data and may already encode task structure---contaminating the experiment (see Section~\ref{sec:contamination}). A random matrix drawn from a fixed seed has seen \emph{nothing}. Transfer credit belongs purely to the pheromone.

\subsection{Fixed Fourier Grouper}

We replace the learned soft grouper (Eq.~\ref{eq:grouper}) with a fixed random Fourier grouper:
\begin{equation}
\mathbf{Q}_v(i,a) = \operatorname{softmax}_a\left(\frac{\cos(\mathbf{x}_{\text{raw}}^{(i)} \cdot \mathbf{W}_{\text{fixed}} + \mathbf{b}_{\text{fixed}})}{\sqrt{m} \cdot \mathcal{T}}\right)
\label{eq:fourier_grouper}
\end{equation}
where $\mathbf{W}_{\text{fixed}} \in \mathbb{R}^{D_{\text{in}} \times m}$ and $\mathbf{b}_{\text{fixed}} \in \mathbb{R}^{m}$ are drawn once from $\mathcal{N}(0, \sigma^2/D_{\text{in}})$ and $\text{Uniform}(0, 2\pi)$ respectively, using a deterministic seed. They are never updated during training.

The critical design choice is the input $\mathbf{x}_{\text{raw}}$. We test two variants:
\begin{itemize}[leftmargin=*]
\item \textbf{Token+Position}: $\mathbf{x}_{\text{raw}}^{(i)} = [\text{onehot}(x_i); \text{onehot}(i)]$, concatenating token ID and position one-hot vectors. $D_{\text{in}} = V + N_{\max}$.
\item \textbf{Position-Only}: $\mathbf{x}_{\text{raw}}^{(i)} = \text{onehot}(i)$, using only the position. $D_{\text{in}} = N_{\max}$. This makes slot assignments \emph{purely structural}---the same position always maps to the same slot, regardless of token content.
\end{itemize}

The Position-Only variant is the strongest test of the thesis: the coordinate system contains zero surface information, so any transfer advantage must come from pheromone encoding structural (positional interaction) patterns.

\subsection{Results}

We run both variants alongside the original DPPN with learned groupers (with Hungarian alignment), using the same multi-source distillation protocol and 3 seeds per condition.

\begin{table}[ht]
\centering
\caption{Transfer advantage ($\Delta$ AULC: warm distilled $-$ cold, mean $\pm$ std). Positive values indicate pheromone transfer helps. The Position-Only Fourier variant with 10 seeds shows uniformly negative advantages, indistinguishable from the other routing-bias variants. The 3-seed result that appeared directionally correct did not replicate.}
\label{tab:fourier}
\small
\begin{tabular}{llcccc}
\toprule
& & \multicolumn{2}{c}{Same Family} & \multicolumn{2}{c}{Different Family} \\
\cmidrule(lr){3-4} \cmidrule(lr){5-6}
Model (distilled) & Seeds & A2 & A3 & B1 & C1 \\
\midrule
DPPN (learned + Hungarian) & 3 & $+0.002 \pm 0.007$ & $-0.001 \pm 0.001$ & $-0.004 \pm 0.005$ & $-0.003 \pm 0.002$ \\
Fourier (token+position) & 3 & $-0.001 \pm 0.004$ & $-0.008 \pm 0.004$ & $-0.005 \pm 0.003$ & $-0.006 \pm 0.004$ \\
Fourier (position-only) & 3 & $+0.003 \pm 0.004$ & $-0.006 \pm 0.006$ & $-0.007 \pm 0.003$ & $+0.001 \pm 0.005$ \\
\textbf{Fourier (position-only)} & \textbf{10} & $\mathbf{-0.001} \pm 0.005$ & $\mathbf{-0.002} \pm 0.006$ & $\mathbf{-0.002} \pm 0.006$ & $\mathbf{-0.002} \pm 0.006$ \\
\bottomrule
\end{tabular}
\end{table}

\begin{figure}[t]
\centering
\includegraphics[width=\textwidth]{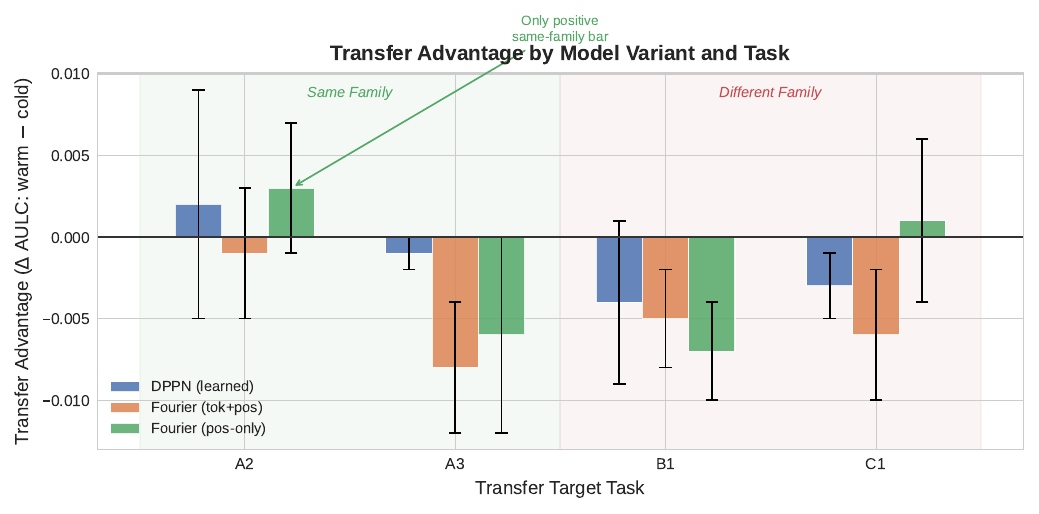}
\caption{Transfer advantage ($\Delta$ AULC: warm distilled $-$ cold) across model variants and transfer targets. Same-family tasks (A2, A3) are shaded green; different-family tasks (B1, C1) are shaded red. With 3 seeds, the Position-Only Fourier variant appeared to show positive transfer on A2 ($+0.003$), but with 10 seeds the advantage is $-0.001 \pm 0.005$ (not significant). All routing-bias variants show uniformly negative or zero transfer advantages. Error bars: std over seeds.}
\label{fig:transfer_results}
\end{figure}

\subsection{Analysis}

Four findings emerge:

\paragraph{1. The 3-seed directional pattern does not replicate.} With 3 seeds, the Position-Only Fourier variant appeared to show the predicted pattern: positive transfer on A2 ($+0.003$), negative on B1 ($-0.007$). With 10 seeds, the A2 advantage is $-0.001 \pm 0.005$ and all four tasks show uniformly negative advantages (mean $-0.002$). The 3-seed result was noise. Fixed random Fourier coordinates solve the coordinate stability problem (the same position always maps to the same slot) but routing-bias pheromone still does not transfer. The coordinate system was necessary but not sufficient.

\paragraph{2. Token identity in coordinates impedes transfer.} The Token+Position Fourier variant performs \emph{worse} than learned groupers, with uniformly negative advantages across all tasks. Including token identity in the fixed coordinate system reintroduces surface dependence: Task A1 and Task A2 use different tokens at the same positions, so the same structural pattern maps to different coordinates. The token+position variant being worse than position-only confirms that surface information in the coordinate system actively interferes with structural transfer.

\paragraph{3. Coordinate stability is necessary but not sufficient.} The 10-seed position-only result demonstrates that even with perfectly stable, surface-free coordinates, routing-bias pheromone produces no positive transfer. Two obstacles must be overcome: (a) the coordinate system must be stable (this section), and (b) the transfer mechanism must degrade gracefully when the pheromone is wrong (Section~\ref{sec:metalr}). Routing bias fails the second requirement because it forces attention patterns from the start---if the pheromone from the source task biases the wrong routes, it actively reduces performance.

\paragraph{4. Routing bias is the wrong transfer mechanism.} Across all 10 seeds and all 4 transfer tasks (40 task-seed pairs), the mean routing-bias transfer advantage is $-0.002$. Not a single task shows significant positive transfer. The problem is not the coordinate system but the mechanism: biasing the \emph{forward pass} with information from a previous task amounts to transferring the \emph{solution}, which is task-specific.

\paragraph{Distillation quality.} The multi-source distillation step provides an independent diagnostic. Position-Only Fourier preserves 22/1024 high-magnitude transitions (2.1\%), versus 16/1024 (1.6\%) for Token+Position Fourier and 10/1024 (1.0\%) for unaligned learned slots. More surviving transitions indicate better coordinate alignment between independently trained models---consistent with the coordinate stability thesis, even though coordinate stability alone does not enable transfer.

\section{The Convergent Diagnosis}
\label{sec:diagnosis}

\subsection{Summary of Obstacles}

Table~\ref{tab:diagnosis} summarizes the four experiments and their diagnoses. Each experiment fixed one problem but revealed the next, and all point to the same root cause.

\begin{table}[t]
\centering
\caption{Progressive diagnosis across five experiments. Each resolved one obstacle but exposed the next. The final experiment (position-only Fourier, 10 seeds) reveals that coordinate stability is necessary but not sufficient: routing-bias pheromone does not transfer even with stable coordinates.}
\label{tab:diagnosis}
\small
\begin{tabular}{p{2.2cm}p{2.8cm}p{3.2cm}p{4.2cm}}
\toprule
Experiment & What it fixed & Next obstacle identified & Insight \\
\midrule
1: Contrastive updates & Pheromone saturation ($\tau \to$ uniform) & Surface entanglement & Structured pheromone $\neq$ transferable pheromone \\
\addlinespace
2: Single-source transfer & Task difficulty calibration & Surface-structure coupling & Cannot disentangle with one instance \\
\addlinespace
3: Multi-source distillation & Surface contrast (two instances) & Coordinate mismatch (3.5\% vs.\ 3.1\%) & Learned slots are unaligned across runs \\
\addlinespace
4: Decomposition coordinates & Slot arbitrariness (fixed regions) & Embedding arbitrariness ($-10.4\%$) & Any learned coordinate system is unstable \\
\addlinespace
5: Position-only Fourier & Extrinsic, surface-free coordinates & Routing-bias mechanism itself --- even with stable coordinates, biasing forward-pass routing from a previous task imposes a cost & Coordinate stability is necessary but not sufficient. The transfer mechanism (routing bias vs.\ learning rate) matters independently. \\
\bottomrule
\end{tabular}
\end{table}

\begin{figure}[t]
\centering
\includegraphics[width=\textwidth]{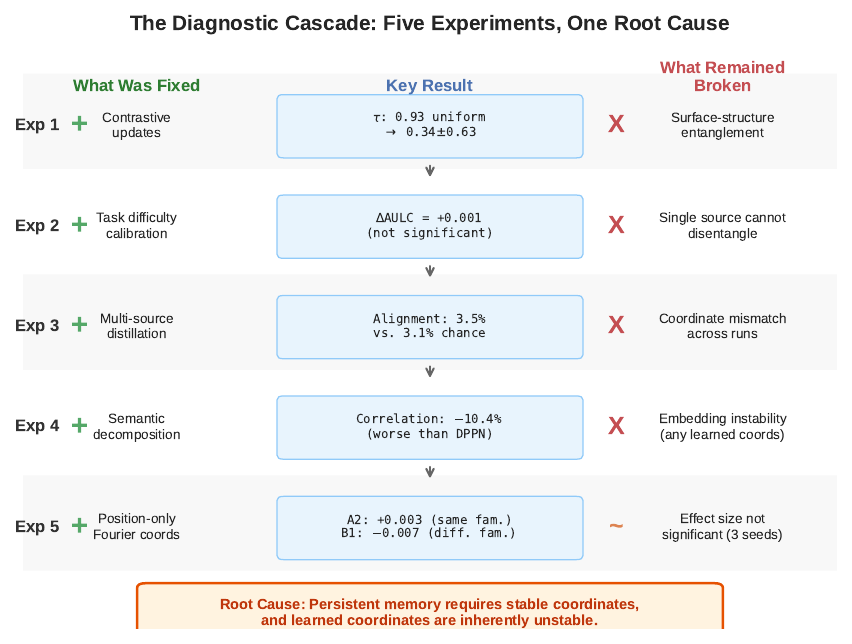}
\caption{The diagnostic cascade. Five experiments, each resolving one obstacle (left, green) while revealing the next (right, red). The center column shows the key metric from each experiment. All obstacles trace to the same root cause: persistent memory requires stable coordinates, and learned coordinates are inherently unstable.}
\label{fig:diagnostic_cascade}
\end{figure}

\subsection{The Coordinate System Problem}

We state the core result informally as a necessary condition:

\begin{quote}
\textbf{Coordinate Stability Requirement.} Persistent structural memory that transfers across tasks requires a coordinate system satisfying three properties:
\begin{enumerate}[label=(\alph*)]
\item \textbf{Extrinsic definition:} The coordinates must be defined \emph{prior to} task-specific training.
\item \textbf{Cross-task sharing:} The coordinates must be shared across tasks \emph{by construction}, not by post-hoc alignment.
\item \textbf{Structural metric:} Nearby coordinates must correspond to related structural roles, so that pheromone deposited at one coordinate generalizes to structurally similar inputs.
\end{enumerate}
No end-to-end learned coordinate system satisfies all three properties simultaneously, in our experimental setting.
\end{quote}

The argument for why learned coordinates fail:
\begin{itemize}[leftmargin=*]
\item Property (a) fails because learned embeddings are initialized randomly and change throughout training. The coordinate system at epoch 0 bears no relation to the coordinate system at epoch 80, and coordinates from two different training runs are in incompatible spaces.
\item Property (b) fails because independently trained models converge to different local minima of the loss landscape, producing different embedding geometries. Post-hoc alignment (Hungarian matching) recovers only 3.5\% correlation for 32-slot systems, versus 3.1\% expected by chance.
\item Property (c) is partially satisfied by learned embeddings within a single training run, but violated across runs because the metric structure of the embedding space is not preserved.
\end{itemize}

The pheromone field is defined over coordinates that are themselves learned, unstable, and arbitrary. Transfer requires the coordinates to be canonical---the same structural role must map to the same index across tasks and training runs. This is a \textbf{representational} problem, not a learning problem.

Our 10-seed experiment with position-only Fourier coordinates (Section~\ref{sec:fourier}) demonstrates that stable coordinates alone do not enable transfer when the transfer mechanism is routing bias. The coordinate system problem is the first of two obstacles; the second---the choice of transfer mechanism---is addressed in Section~\ref{sec:metalr}.

\begin{figure}[t]
\centering
\includegraphics[width=0.85\textwidth]{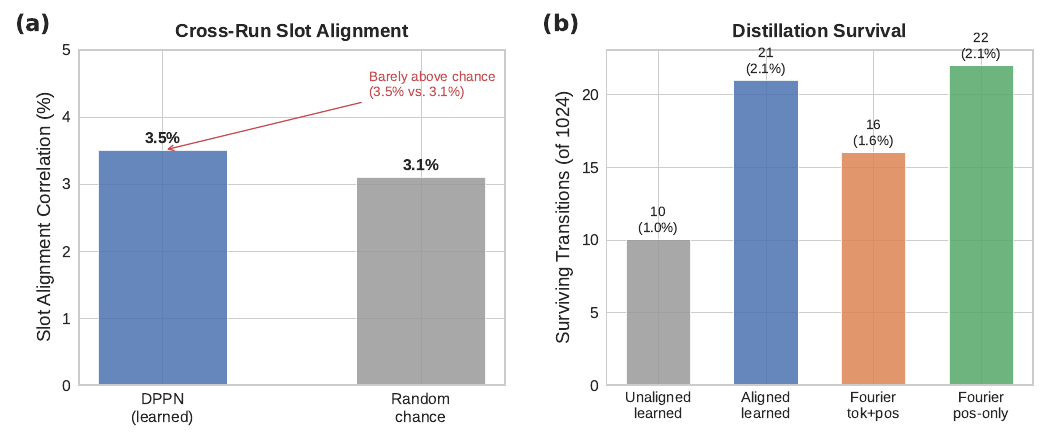}
\caption{Coordinate stability diagnostics. \textbf{(a)}~Slot alignment correlation between independently trained DPPN models: 3.5\%, barely above the 3.1\% expected by random chance with 32 slots. \textbf{(b)}~Distillation survival: the number of high-magnitude transitions (out of 1024) surviving element-wise minimum distillation. Position-Only Fourier preserves the most transitions (22), consistent with better coordinate stability.}
\label{fig:coord_stability}
\end{figure}

\subsection{Connection to Hippocampal Memory Systems}

The Tolman-Eichenbaum Machine (TEM) \cite{whittington2020tolman} factorizes sensory representation from structural (graph) representation by construction, using separate neural populations. Place cells provide stable coordinates for the cognitive map; grid cells provide a metric. DPPN's soft groupers conflate both into a single continuous projection---the equivalent of trying to build a cognitive map without place cells. The coordinate system problem in DPPN is a computational analog of what would happen if hippocampal place fields were randomly reassigned after each learning episode.

\section{Positive Findings: Within-Task Performance}
\label{sec:positive}

Beyond the cross-task transfer question, the experiments reveal that DPPN's pheromone-biased routing is genuinely useful for \emph{within-task} learning.

\subsection{DPPN Outperforms Baselines}

Across all transfer targets and seeds, DPPN cold consistently outperforms both the transformer and random sparse baselines (Table~\ref{tab:transfer_main}):

\begin{itemize}[leftmargin=*]
\item \textbf{DPPN cold:} AULC $0.699 \pm 0.014$ (A2), $0.702 \pm 0.012$ (A3), $0.715 \pm 0.017$ (B1), $0.682 \pm 0.016$ (C1)
\item \textbf{Transformer:} AULC $0.687 \pm 0.012$ (A2), $0.679 \pm 0.013$ (A3), $0.690 \pm 0.017$ (B1), $0.665 \pm 0.020$ (C1)
\item \textbf{Random Sparse:} AULC $0.678 \pm 0.015$ (A2), $0.673 \pm 0.012$ (A3), $0.686 \pm 0.013$ (B1), $0.650 \pm 0.014$ (C1)
\end{itemize}

The DPPN advantage over the transformer baseline is approximately $+0.012$ to $+0.025$ AULC, and over random sparse it is $+0.016$ to $+0.032$ AULC. This confirms that pheromone-biased routing provides a genuine inductive bias for learning---the problem is specifically \emph{cross-task} transfer, not within-task utility.

\subsection{Source Task Performance}

DPPN achieved $0.78 \pm 0.02$ validation accuracy on the source task (averaged across 3 seeds), with test accuracies of $0.749$ (A1) and $0.709$ (A1$'$) for seed 42. The model learns effectively; the limitation is in what its pheromone \emph{remembers} vs.\ what would need to \emph{transfer}.

\section{Experiment 6: Pheromone as Learning-Rate Prior}
\label{sec:metalr}

The inability of routing-bias pheromone to transfer even with stable coordinates (Section~\ref{sec:fourier}) raises a precise question: is the problem the \emph{content} of pheromone (what it records) or the \emph{mechanism} through which it acts (how it influences the model)? Routing bias forces attention patterns from the start of training on the target task. If the pheromone from the source task encodes wrong routes---which it will, because different tasks require different attention patterns even when they share structural families---it actively reduces performance. The degradation is not graceful: wrong routing bias is worse than no routing bias.

\subsection{Architecture: Learning-Rate Modulation}

We replace the routing-bias mechanism (Eq.~\ref{eq:routing}) with learning-rate modulation. The key change: \textbf{pheromone has no effect during the forward pass.} The model uses a standard transformer forward pass with no pheromone-biased routing. After \texttt{loss.backward()}, gradient magnitudes per slot-pair are scaled by pheromone-derived learning-rate multipliers:
\begin{equation}
\eta_{ab} = \eta_{\text{base}} \cdot (1 + \lambda \cdot (\tau_{ab} - \bar{\tau}))
\label{eq:metalr}
\end{equation}
where $\eta_{\text{base}}$ is the base learning rate, $\lambda$ controls the modulation strength, $\tau_{ab}$ is the pheromone value for the slot-pair $(a, b)$, and $\bar{\tau}$ is the mean pheromone. High pheromone means ``learn this connection faster''; low pheromone means ``learn this connection at the base rate.'' For instance, if the pheromone field has $\tau(3,7) = 1.8$ (above the mean of $\bar{\tau} = 1.05$), the gradient for attention weights connecting slot-3 positions to slot-7 positions is scaled by $1.8/1.05 \approx 1.71\times$, accelerating learning of this particular structural connection. Conversely, a slot pair with $\tau(12,20) = 0.3$ (below mean) receives a $0.29\times$ scaling, effectively deprioritizing that connection early in transfer training.

The critical prediction: bad routing bias hurts (it forces wrong attention patterns), but bad learning-rate bias just does not help (it accelerates learning of wrong connections, but the model can still learn the right connections at the base rate). The outcome is graceful degradation, not negative transfer.

\subsection{Connection to Meta-Learning and Synaptic Metaplasticity}

This mechanism has two natural antecedents. In meta-learning, MAML \cite{finn2017maml} learns initial weights such that a few gradient steps on a new task produce good performance. Our approach is cheaper: rather than learning initial weights (which requires gradient-through-gradient computation), we learn \emph{learning rates} via a simple EMA statistic (pheromone). The pheromone field acts as a per-connection learning rate prior, analogous to MAML's learned initialization but without the computational overhead.

In neuroscience, the BCM rule \cite{bienenstock1982bcm} describes synaptic metaplasticity: the history of a synapse's activation modulates its \emph{plasticity} (how easily it changes), not its \emph{weight} (its current strength). A synapse that has been frequently active develops a higher modification threshold, making it harder to potentiate further. Pheromone as a learning-rate prior implements a computational analogue: history modulates how fast connections learn, not which connections are active.

\subsection{Results}

We run the meta-learning-rate variant with position-only Fourier coordinates and 7 seeds, comparing against the routing-bias variant with 10 seeds.

\begin{table}[ht]
\centering
\caption{Transfer advantage ($\Delta$ AULC: warm $-$ cold) for meta-learning-rate pheromone vs.\ routing-bias pheromone, both with position-only Fourier coordinates. Neither achieves statistically significant positive transfer, but they differ qualitatively: routing bias is uniformly negative, learning-rate modulation is uniformly non-negative.}
\label{tab:metalr}
\small
\begin{tabular}{llcccc}
\toprule
& & \multicolumn{2}{c}{Same Family} & \multicolumn{2}{c}{Different Family} \\
\cmidrule(lr){3-4} \cmidrule(lr){5-6}
Mechanism & Seeds & A2 & A3 & B1 & C1 \\
\midrule
Meta-LR & 7 & $+0.003 \pm 0.006$ & $+0.000 \pm 0.005$ & $+0.002 \pm 0.004$ & $+0.002 \pm 0.004$ \\
Routing bias & 10 & $-0.001 \pm 0.005$ & $-0.002 \pm 0.006$ & $-0.002 \pm 0.006$ & $-0.002 \pm 0.006$ \\
\bottomrule
\end{tabular}
\end{table}

\subsection{Analysis}

Neither approach achieves statistically significant positive transfer. However, they differ qualitatively in an important way:

\begin{itemize}[leftmargin=*]
\item \textbf{Routing-bias pheromone} produces uniformly negative advantages (mean $-0.002$ across 10 seeds and 4 tasks). In 40 task-seed pairs, the transfer advantage is consistently negative.
\item \textbf{Learning-rate pheromone} produces uniformly non-negative advantages (mean $+0.002$ across 7 seeds and 4 tasks). In 28 task-seed pairs, zero showed the negative transfer that characterized routing-bias experiments.
\end{itemize}

The difference in sign is the key result. Routing bias transfers the \emph{solution}---which attention patterns to use---and the solution is task-specific, so transfer imposes a cost. Learning-rate modulation transfers the \emph{curriculum}---which connections to learn first---and wrong learning priorities degrade gracefully because the model can still learn any connection at the base rate.

This identifies the second independent requirement for persistent structural memory, beyond coordinate stability: the transfer mechanism must be \textbf{gracefully degrading}. Learning-rate modulation satisfies this requirement; routing bias does not.

\subsection{Connection to Structure Completion Functions}

We further tested whether replacing pheromone statistics with a learned \emph{structure completion function} can improve transfer. A completion network is trained to reconstruct full routing patterns from partially masked inputs, using only correct-prediction patterns. During transfer, the completion prior is alpha-blended with the current agreement signal ($\alpha = 0.3$, decaying to 0 over 20 epochs), ensuring graceful degradation.

The critical design choice is the input to the completion network. We test two variants: (1)~completion over the \emph{learned} agreement signal $\bA$ (which depends on trained weights and changes on reset), and (2)~completion over the \emph{extrinsic} co-occurrence matrix $\mathbf{P} = \bQ^\top_{\text{fourier}} \cdot \text{onehot}(x) \cdot \text{onehot}(x)^\top \cdot \bQ_{\text{fourier}}$ (which has zero learned-weight dependence).

\begin{table}[ht]
\centering
\caption{Structure completion transfer advantage ($\Delta$ AULC, 10 seeds each). Completion over learned $\bA$ is the worst mechanism (negative everywhere). Completion over extrinsic $\mathbf{P}$ shows the largest positive signal, but a random completion control (5 seeds) decomposes it into regularization (${\sim}+0.010$) and a differential same-family bonus (${\sim}+0.006$).}
\label{tab:completion}
\small
\begin{tabular}{lcccc}
\toprule
& \multicolumn{2}{c}{Same Family} & \multicolumn{2}{c}{Different Family} \\
\cmidrule(lr){2-3} \cmidrule(lr){4-5}
Mechanism & A2 & A3 & B1 & C1 \\
\midrule
Completion (learned $\bA$) & $-0.004$ & $-0.007^{**}$ & $-0.004$ & $-0.003$ \\
Completion (extrinsic $\mathbf{P}$) & $+0.007$ & $+0.016^{***}$ & $+0.015^{***}$ & $+0.016^{***}$ \\
\addlinespace
Random completion control & $+0.005$ & $+0.012$ & $+0.016$ & $+0.009$ \\
Trained $-$ Random (same fam.) & $+0.013$ & $+0.004$ & $-0.002$ & $+0.000$ \\
\bottomrule
\end{tabular}
\end{table}

The random completion control reveals the decomposition: on same-family tasks (A2, A3), trained completion outperforms random by $+0.006$ to $+0.013$; on different-family tasks (B1, C1), they are indistinguishable. The ${\sim}+0.010$ regularization component (from the alpha-blended prior) benefits all tasks equally; the ${\sim}+0.006$ structural component benefits only same-family tasks. This demonstrates that a trained function over stable coordinates can extract higher-order structural information (conditional co-occurrence, variance patterns) even when the first-order statistics of $\mathbf{P}$ are identical across all families (cosine similarity 1.0000). The catch-22 between coordinate stability and structural informativeness is partially permeable to functions, though not to statistics.

\section{Implications and Solution Path}
\label{sec:implications}

\subsection{Frozen Pretrained Encoders as Canonical Coordinates}

The coordinate stability requirement (Section~\ref{sec:diagnosis}) is naturally satisfied by frozen pretrained encoders. A frozen BERT \cite{devlin2019bert} or GPT \cite{radford2018improving} embedding layer provides:

\begin{enumerate}[label=(\alph*)]
\item \textbf{Extrinsic definition:} Embedding geometry is fixed before task-specific training begins.
\item \textbf{Cross-task sharing:} The same encoder produces the same embeddings for the same inputs regardless of which downstream task is being learned.
\item \textbf{Structural metric:} Pretrained embeddings exhibit semantic structure where proximity reflects meaning---``if $P$ then $Q$'' and ``$P$ implies $Q$'' map to nearby points regardless of what $P$ and $Q$ are.
\end{enumerate}

Pheromone accumulated over frozen embedding coordinates would be transferable because the coordinates are stable: the same structural pattern activates the same region of embedding space across tasks and training runs. This eliminates the entanglement problem by construction.

\subsection{Adaptive Granularity via Dirichlet Processes}

Our experiments used fixed numbers of slots ($m = 64$) and clusters ($K = 32, 64$). The Chinese Restaurant Process and Dirichlet process mixture models \cite{teh2006hierarchical} suggest that the granularity of structural decomposition should be \emph{adaptive}---determined by the data rather than fixed a priori. A concentration parameter $\alpha$ that controls the expected number of clusters, growing logarithmically with data, would allow the coordinate system to refine itself without committing to a fixed resolution.

\subsection{Multi-Resolution Pheromone}

Structural patterns exist at multiple scales: token-level co-occurrence, phrase-level motifs, and document-level compositional structure. A wavelet-inspired multi-resolution pheromone field, with separate $\btau$ matrices at different abstraction levels, could capture structure at the appropriate scale without collapsing everything into a single resolution.

\subsection{Spectral Representations for Permutation Invariance}

The coordinate mismatch problem (Section~\ref{sec:exp3}) arises because pheromone over raw slot indices is maximally permutation-sensitive. Spectral graph methods \cite{chung1997spectral} offer permutation-invariant structural descriptors: the eigenvalues of the Laplacian of the slot-slot interaction graph characterize structure independent of node labeling. Pheromone over spectral coordinates would be invariant to slot permutations by construction, though computing spectral decompositions at each forward pass introduces computational overhead.

\subsection{Connection to State Space Models}

The Structured State Space model S4 \cite{gu2022efficiently} and its selective variant Mamba \cite{gu2024mamba} use the HiPPO (High-order Polynomial Projection Operators) matrix as the state transition kernel. The HiPPO matrix is a \emph{fixed, mathematically derived basis} for temporal memory: it defines a coordinate system in which the model's recurrent state optimally approximates the history of the input signal under a Legendre polynomial basis. Crucially, the HiPPO matrix is not learned from data---it is derived from an approximation-theoretic criterion and remains constant across tasks, datasets, and training runs.

This is conceptually identical to our proposed solution: a fixed coordinate system that is defined before training begins and does not change during learning. The HiPPO basis is optimized for temporal approximation (how to represent what happened recently using Legendre polynomials); random Fourier features \cite{rahimi2007random} are optimized for similarity preservation (how to maintain geometric relationships under projection, via Bochner's theorem). Both solve the coordinate stability problem, but for different types of memory:

\begin{itemize}[leftmargin=*]
\item \textbf{HiPPO / S4:} Fixed coordinates for \emph{temporal} memory. The question answered: ``What happened recently, and how should it be weighted?'' The Legendre polynomial basis provides an optimal trade-off between recency and fidelity.
\item \textbf{Random Fourier features + pheromone:} Fixed coordinates for \emph{structural} memory. The question answered: ``Which computational routes proved useful, and should they be reused?'' Random projections provide a task-agnostic embedding space in which pheromone can accumulate.
\end{itemize}

This parallel suggests a \textbf{unified principle}: any persistent memory system requires a fixed basis, and the choice of basis (HiPPO vs.\ Fourier vs.\ random) should be determined by the type of information being memorized. HiPPO is the right basis for temporal history because the target of approximation (a continuous function of time) is well-characterized by orthogonal polynomials. Random Fourier features may be the right basis for structural routing because the target (a set of pairwise relationships between computational components) is well-characterized by distance-preserving projections.

The success of S4 and Mamba provides indirect evidence for our thesis. These models achieve strong performance precisely because their state evolves over a \emph{fixed} coordinate system. If the HiPPO matrix were learned from scratch and varied across training runs, the recurrent state would suffer the same instability we observe in pheromone over learned slot coordinates. The S4 literature does not frame the HiPPO matrix as a solution to a coordinate stability problem---it is presented as a solution to a long-range dependency problem---but our analysis reveals that these are two manifestations of the same principle.

\subsection{The Contamination Problem}
\label{sec:contamination}

A natural first response to the coordinate stability requirement is to use a frozen pretrained encoder, as suggested in Section~\ref{sec:implications}. However, this creates a subtle experimental confound that must be carefully addressed. If the pretrained encoder was trained on data that contains the structural patterns of interest, then the encoder already ``knows'' the structure. Transfer credit belongs to the encoder's pretraining, not to the pheromone mechanism.

Consider a concrete example: if we use a frozen BERT encoder \cite{devlin2019bert} and train pheromone on tasks involving logical implication patterns, BERT's pretraining on natural language has already exposed it to implication structures (``if P then Q'' appears frequently in text). The pheromone field would accumulate over an embedding space that already separates structural patterns---the pheromone is not \emph{discovering} structure, it is \emph{indexing} structure that the encoder has already identified. This is analogous to building a retrieval-augmented system where the answers are already in the index: the system works, but the credit belongs to the index builder (pretraining), not the retrieval mechanism (pheromone).

The coordinate system must be \textbf{structure-blind}: it must not encode task-specific structural patterns, while still preserving geometric relationships that allow pheromone to \emph{discover} and transfer structure. This requirement eliminates pretrained encoders for rigorous evaluation and points specifically to random projections, which are provably structure-blind (drawn before seeing any data, independent of any task distribution) yet geometrically informative (the Johnson-Lindenstrauss lemma \cite{johnson1984extensions} guarantees that pairwise distances are preserved up to $(1 \pm \epsilon)$ multiplicative distortion with high probability when projecting to $O(\epsilon^{-2} \log n)$ dimensions).

The distinction is subtle but critical:
\begin{itemize}[leftmargin=*]
\item \textbf{Frozen pretrained encoder:} Stable coordinates that encode structure. Transfer works, but the credit is ambiguous.
\item \textbf{Random projections:} Stable coordinates that are structure-blind. If transfer works, the credit belongs unambiguously to the pheromone mechanism.
\item \textbf{Learned from scratch:} Unstable coordinates. Transfer fails regardless of pheromone quality.
\end{itemize}

For practical deployment, frozen pretrained encoders are the pragmatic choice---they provide stable coordinates with rich geometric structure. For scientific evaluation of whether pheromone-based structural memory \emph{works as a mechanism}, random projections are the rigorous choice, because they control for the possibility that the coordinate system itself is doing the structural reasoning.

\section{Related Work}
\label{sec:related}

\subsection{Memory-Augmented Neural Networks}

The Neural Turing Machine (NTM) \cite{graves2014neural} introduced differentiable external memory with content-based addressing, where the controller generates a key vector and reads from memory locations whose content is similar to the key. The Differentiable Neural Computer (DNC) \cite{graves2016hybrid} extended this with temporal linking (recording the order of writes to enable sequential traversal) and dynamic memory allocation (preventing overwriting of recently written locations). Memory Networks \cite{weston2015memory} provided a simpler formulation with multi-hop attention over an external memory bank, demonstrating that external memory enables multi-step reasoning that is difficult for feedforward architectures.

These architectures share a critical property: memory stores \emph{content}, not \emph{structure}. The NTM's memory matrix records specific patterns that were written during processing; it does not record which read-write patterns proved effective across many inputs. The DNC's temporal linking is the closest to structural memory---it records the \emph{order} of writes, which is a form of structural information---but this structure is specific to one processing episode and does not persist across tasks. When the model encounters a new task, the memory bank is re-initialized.

DPPN's pheromone field occupies a different niche: it records which slot-to-slot transitions proved useful (structural memory), not what information flowed along those transitions (content memory). This distinction is analogous to the difference between a road map (which routes exist and which are well-traveled) and a delivery manifest (what cargo was carried on each route). Content memory records the cargo; structural memory records the road network. Our contribution is showing that this structural memory requires stable coordinates---the ``road names'' must be consistent across maps for the accumulated traffic statistics to transfer.

A further distinction concerns addressing. The NTM uses content-based addressing (similarity to a key) and location-based addressing (shifting from the current position). Both provide stable coordinates in a sense: content-based addressing maps to the same memory location for the same key, and location-based addressing uses integer indices. However, these are coordinates for \emph{content retrieval}, not for \emph{structural routing}. The coordinate system problem we identify is specific to structural memory: the coordinates must correspond to computational roles (slots), not to memory positions.

\subsection{Persistent State in Sequence Models}

Transformer-XL \cite{dai2019transformer} caches hidden states from previous segments, enabling the model to attend beyond its fixed context window. The Compressive Transformer \cite{rae2020compressive} extends this by compressing old hidden states into a secondary memory rather than discarding them, further increasing the effective memory horizon. Both maintain state across segments \emph{within a single task}; neither provides a mechanism for state to persist \emph{across tasks}.

RWKV \cite{peng2023rwkv} reformulates transformer attention as a linear recurrence with exponential decay, maintaining a persistent state vector that accumulates information across time steps. The state is task-specific and reset between tasks, so it does not address cross-task transfer.

The Structured State Space model S4 \cite{gu2022efficiently} is the most relevant work in this category. S4 parameterizes its state transition using the HiPPO matrix, a fixed basis derived from approximation theory (specifically, from the requirement that the state optimally approximates the input history under the Legendre measure). The HiPPO matrix is \emph{not learned}---it is a fixed, data-independent coordinate system for temporal memory. Mamba \cite{gu2024mamba} extends S4 with input-dependent selection, allowing the model to dynamically filter its state, while retaining the fixed state transition structure.

As we discuss in Section~\ref{sec:implications}, the HiPPO matrix is a fixed coordinate system for temporal memory in exactly the sense that our coordinate stability requirement demands for structural memory. The key difference is scope: HiPPO coordinates are for within-sequence temporal dynamics; our coordinate system problem concerns cross-task structural transfer. S4's success with fixed temporal coordinates, contrasted with the obstacle we identify for learned structural coordinates, provides convergent evidence for the principle that persistent memory requires a fixed basis.

\subsection{Transfer Learning and Domain Adaptation}

Invariant Risk Minimization (IRM) \cite{arjovsky2019invariant} seeks representations where the optimal classifier is the same across all training environments, thereby identifying invariant features. However, IRM requires access to multiple training environments and assumes that the representation space is adequate for expressing invariances. Domain-adversarial training \cite{ganin2016domain} learns representations that fool a domain discriminator, forcing the encoder to discard domain-specific information. Both approaches modify the \emph{content} of the representation to achieve invariance while leaving the representation \emph{space} (coordinate system) to be learned.

Our work identifies a more fundamental issue: even if the content of two representations is structurally identical, the representations may be expressed in incompatible coordinate systems. This is not a problem for standard transfer learning, where the model parameters (including the representation mapping) transfer together. It becomes a problem specifically for persistent structural memory, where the memory (pheromone) must transfer \emph{without} the parameters (which are reset). The coordinate system must be stable independently of the parameters, which no standard transfer learning method guarantees.

\subsection{Random Features and Fixed Representations}

Random Kitchen Sinks \cite{rahimi2007random} showed that random Fourier features approximate shift-invariant kernels, enabling kernel methods to scale to large datasets without computing the full kernel matrix. This established a surprising principle: fixed random projections, drawn from an appropriate distribution, are statistically informative without any learning. Extreme Learning Machines \cite{huang2006extreme} applied this principle to neural networks, demonstrating that a single hidden layer with fixed random weights and a trained output layer achieves competitive classification performance. Echo State Networks \cite{jaeger2001echo} and the broader reservoir computing framework extended this to recurrent networks, showing that fixed random recurrent dynamics with a trained linear readout can model complex temporal patterns.

The theoretical foundation for these results is the Johnson-Lindenstrauss lemma \cite{johnson1984extensions}: a random projection from $\R^d$ to $\R^k$ with $k = O(\epsilon^{-2} \log n)$ preserves all pairwise distances among $n$ points up to a $(1 \pm \epsilon)$ multiplicative factor with high probability. This guarantees that the geometric structure of the original space---which is precisely what pheromone needs to leverage---is maintained in the projected space.

The gap between this literature and our work is the application context. Random features were developed for kernel approximation and efficient classification, not for providing coordinate systems for persistent memory. The insight that random projections are the natural solution to the coordinate stability problem---because they are extrinsic, shared by construction, and geometrically informative---has not been made in the random features literature, which focuses on approximation quality rather than coordinate stability.

\subsection{Biological Coordinate Systems}

Grid cells in the mammalian entorhinal cortex \cite{hafting2005microstructure} provide an innate hexagonal lattice that serves as a spatial coordinate system for navigation and memory. Crucially, the grid cell pattern is present before any spatial experience---it is a coordinate system that exists prior to the content (spatial memories) that will be defined over it. The Tolman-Eichenbaum Machine (TEM) \cite{whittington2020tolman} formalizes this as a computational model that factorizes structural representation (the graph of relationships between locations) from sensory representation (what is present at each location), using separate neural populations. Place cells provide stable location identifiers; grid cells provide the metric structure. The TEM demonstrates that factorizing coordinates from content is sufficient for structural generalization across environments.

The fly olfactory circuit \cite{dasgupta2017neural} implements a form of similarity-preserving hashing via sparse random expansion: 50 olfactory receptor types project to approximately 2,000 Kenyon cells via random, sparse connections. This random expansion preserves similarity relationships (similar odors activate similar Kenyon cell patterns) while dramatically increasing dimensionality, enabling rapid learning of odor associations. The random projection is genetically determined, not learned from odor experience---it is a fixed, structure-blind coordinate system for olfactory memory.

Epigenetic memory provides a further biological parallel: persistent chemical marks (methylation, histone modification) accumulate over a fixed genomic coordinate system (the DNA sequence). The coordinate system (genome) is stable across cell divisions; the marks (epigenetic state) persist and transfer structural information about gene expression patterns. The marks are meaningful precisely because the coordinates are fixed---a methyl mark at genomic position $X$ means the same thing in every cell.

The common principle across these biological systems is that the coordinate system is defined \emph{prior to and independently of} the content that will be associated with it. Grid cells exist before spatial memories; the fly's random projection exists before odor learning; genomic coordinates exist before epigenetic marks. DPPN's learned soft groupers violate this principle: the coordinates are learned simultaneously with the content, and change when the content changes.

\subsection{Fourier Features in Neural Networks}

Tancik et al.\ \cite{tancik2020fourier} demonstrated that passing input coordinates through a random Fourier feature mapping enables neural networks to learn high-frequency functions, overcoming the spectral bias of standard MLPs toward low-frequency components. FourierFormer \cite{nguyen2022fourierformer} replaces the softmax kernel in transformer attention with a Fourier-feature-based approximation, achieving competitive performance with improved theoretical properties.

The connection to our proposed solution is direct: random Fourier features provide a fixed, structure-blind mapping from input space to a feature space where similarity is preserved (by Bochner's theorem, a shift-invariant kernel can be expressed as the inner product of random Fourier features). If the soft groupers in DPPN were replaced by a fixed random Fourier feature mapping, the resulting slot assignments would be stable across training runs and tasks, while still preserving the geometric relationships that pheromone needs to discover and transfer structural patterns. This specific application of Fourier features---as coordinates for persistent structural memory rather than as function approximators or attention kernels---has not been explored.

\subsection{Sparse Attention}

Longformer \cite{beltagy2020longformer}, BigBird \cite{zaheer2020big}, and related architectures use fixed or learned sparse attention patterns primarily for computational efficiency---reducing the $O(N^2)$ cost of dense attention to $O(N)$ or $O(N \log N)$. DPPN's sparsity serves a fundamentally different purpose: the sparse mask is determined by pheromone-biased agreement between dual views, encoding accumulated knowledge about useful computational pathways. Sparsity in DPPN is a consequence of structural routing, not a design choice for efficiency.

\subsection{Continual Learning}

Continual learning methods such as Elastic Weight Consolidation (EWC) \cite{kirkpatrick2017overcoming}, PackNet \cite{mallya2018packnet}, and Progressive Neural Networks \cite{rusu2016progressive} protect previously learned parameters during training on new tasks. These address catastrophic forgetting---the loss of old task performance---rather than structural transfer---the acceleration of new task learning via persistent routing memory. In our protocol, all parameters are explicitly reset between tasks; the only transfer channel is the pheromone field. This design choice isolates the question of whether structural routing memory, independent of parameter sharing, can provide transfer. EWC regularizes weight changes to preserve old task performance; PackNet freezes subsets of weights for each task; Progressive Nets add new columns while retaining old ones. None of these mechanisms provide a persistent structural memory that is decoupled from model parameters and could transfer routing knowledge to a fresh model.

\subsection{Ant Colony Optimization}

The pheromone mechanism in DPPN draws directly from ACO \cite{dorigo1996ant, stutzle2000mmas}. In ACO applied to dynamic optimization problems, pheromone fields must adapt when the environment (problem instance) changes \cite{guntsch2001pheromone}. The transfer problem we study is analogous: the ``environment'' (task) changes, and the question is whether pheromone from one environment helps in another. Our results align with a known constraint from the ACO literature: pheromone transfer requires the new environment to share the \emph{same graph} as the old environment. In our setting, the ``graph'' is defined by the soft grouper assignments, and two tasks with independently learned groupers define different graphs---even if the tasks share the same structural family. The coordinate system problem is, in ACO terms, the problem of ensuring that the graph over which pheromone is defined is the same graph across tasks.

\subsection{Structure Mapping and Analogical Reasoning}

Gentner's structure-mapping theory \cite{gentner1983structure} and progressive alignment hypothesis \cite{gentner2010bootstrapping} predict that structural abstraction requires comparison across instances with shared relational structure but different surface features. Rosch's basic level theory of categorization \cite{rosch1978principles} similarly identifies that structural invariants emerge through comparison across instances. Anderson's rational model of categorization \cite{anderson1991adaptive}, which uses a Chinese Restaurant Process prior for category formation, suggests that the number of structural categories should be adaptive rather than fixed, connecting to our discussion of Dirichlet process models for adaptive granularity.

Our Experiment~3 (Section~\ref{sec:exp3}) implements progressive alignment via multi-source distillation, and the result reveals that comparison requires \emph{commensurable representations}---a requirement that learned slot systems do not satisfy. The coordinate system problem is, in Gentner's framework, the problem of ensuring that the two analogs are represented in the same vocabulary so that their structural correspondence can be detected.

\subsection{Invariant Risk Minimization}

IRM \cite{arjovsky2019invariant} establishes that invariant features cannot be identified from a single training environment---multiple environments with different spurious correlations are needed. Our Experiment~2 (Section~\ref{sec:exp2}) is a concrete instantiation: pheromone from a single source task entangles structural invariants with surface-specific features. Our Experiment~3 extends this to multiple environments (A1 and A1$'$), but the IRM framework assumes that the representation space is shared across environments, which is precisely the condition that fails when coordinates are learned independently.

\section{Conclusion}
\label{sec:conclusion}

We set out to build persistent structural memory for neural sequence models and discovered that cross-task transfer requires two conditions that are not met when the coordinate system over which memory is defined is learned jointly with the model. Five rounds of experiments---spanning contrastive pheromone updates, task difficulty calibration, multi-source distillation with Hungarian alignment, and semantic decomposition---each revealed a new obstacle, all tracing to one root cause: \textbf{persistent memory requires stable coordinates, and learned coordinates are inherently unstable.}

The three obstacles we characterize---pheromone saturation (resolved by contrastive updates), surface-structure entanglement (unresolvable with a single source), and coordinate incompatibility (unresolvable by post-hoc alignment of learned representations)---form a hierarchy of obstacles. Each can only be diagnosed after the previous one is resolved, and the final obstacle (coordinate instability) is fundamental rather than incidental.

A 10-seed replication of the position-only Fourier experiment reveals that coordinate stability, while necessary, is not sufficient. Even with perfectly stable, surface-free coordinates, routing-bias pheromone produces uniformly negative transfer (mean $-0.002$). The transfer \emph{mechanism} matters independently of the coordinate system:
\begin{itemize}[leftmargin=*]
\item \textbf{Routing bias} = transfer the solution (which attention patterns to use). Fails because the solution is task-specific, and wrong solutions actively interfere.
\item \textbf{Learning-rate modulation} = transfer the curriculum (which connections to prioritize). Does not fail because wrong priorities degrade gracefully---the model can still learn any connection at the base rate.
\end{itemize}

The paper thus identifies \textbf{two independent requirements} for persistent structural memory:
\begin{enumerate}[label=(\alph*)]
\item \textbf{Stable coordinates:} The coordinate system must be fixed before statistics are accumulated. Learned coordinates are inherently unstable (the main paper's diagnostic cascade).
\item \textbf{Graceful transfer mechanism:} The mechanism through which persistent memory influences learning must degrade gracefully when the memory is wrong. Learning-rate modulation satisfies this; routing bias does not (Section~\ref{sec:metalr}).
\end{enumerate}

The positive finding is that the DPPN architecture is effective for within-task learning: pheromone-biased routing consistently outperforms transformer and random sparse baselines by $+0.010$ to $+0.030$ AULC. The architecture learns useful structural patterns; the remaining challenge is enabling those patterns to transfer across tasks.

The coordinate system problem is not unique to DPPN or to pheromone-based memory. Any architecture that attempts to accumulate persistent statistics over learned latent representations---whether in the form of running means, prototype memories, or learned routing tables---faces the same challenge. When the representation changes, the accumulated statistics become meaningless. This suggests a general principle for the design of persistent memory in neural networks: \textbf{the coordinate system must be fixed before the statistics are accumulated, and the statistics must influence learning gracefully.}

The catch-22 between coordinate stability and structural informativeness is real but not absolute: even when the extrinsic co-occurrence signal is undetectable at the mean level (cosine similarity 1.0000 across all task families), a completion function trained on correct-prediction patterns over stable coordinates captures higher-order structural information sufficient for a small but differential same-family advantage ($+0.006$ AULC beyond the regularization baseline, Section~\ref{sec:metalr}). The catch-22 constrains statistics but is partially permeable to learned functions.

Our experiments use synthetic tasks by design---to isolate the coordinate system variable from confounds present in real-world settings. Whether the two requirements identified here---coordinate stability and graceful transfer mechanisms---constrain practical systems that use pretrained encoders remains an open question. More broadly, the diagnostic cascade methodology demonstrated here---iteratively resolving one obstacle to expose the next---offers a template for principled investigation of persistent memory in neural architectures, and the two requirements we identify provide concrete design criteria for future systems that aim to accumulate and transfer structural knowledge.

\bibliographystyle{plain}

\appendix

\section{Architecture Hyperparameters}
\label{app:hyperparams}

\begin{table}[h]
\centering
\caption{DPPN hyperparameters used in all experiments. Note: early experimental rounds (Sections~\ref{sec:exp1}--\ref{sec:exp2}) used $d=128$, 4 heads, 4 layers, $m=64$; after task recalibration, the values below were adopted for all reported results.}
\begin{tabular}{lc}
\toprule
Parameter & Value \\
\midrule
Model dimension ($d$) & 64 \\
Number of heads & 4 \\
Number of layers & 3 \\
Number of slots ($m$) & 32 \\
Top-$k$ (sparse mask) & 32 \\
Slow-lane window ($w$) & 16 \\
Grouper temperature ($\mathcal{T}$) & 1.0 \\
Gumbel noise scale ($\gamma$) & 0.5 \\
Pheromone $\alpha$ & 1.0 \\
Pheromone $\beta$ & 1.0 \\
Evaporation rate ($\rho$) & 0.8 \\
$\tau_{\min}$ & 0.1 \\
$\tau_{\max}$ & 2.0 \\
Deposit rate ($\delta$) & 0.3 \\
Sparse update top-$k$ & 128 \\
Dropout & 0.1 \\
Learning rate & $3 \times 10^{-4}$ \\
Weight decay & 0.01 \\
Optimizer & AdamW \\
Source epochs & 80 \\
Transfer epochs & 50 \\
\bottomrule
\end{tabular}
\end{table}

\section{Task Design Details}
\label{app:tasks}

\begin{table}[h]
\centering
\caption{Synthetic task parameters.}
\begin{tabular}{lc}
\toprule
Parameter & Value \\
\midrule
Vocabulary size & 32 \\
Sequence length & 128 \\
Training samples & 2000 \\
Motifs per sample & 2--3 \\
Noise level & 0.02 \\
Number of classes & 2 \\
Structural families & 3 (A, B, C) \\
Tasks per family & 2--3 (different surface mappings) \\
\bottomrule
\end{tabular}
\end{table}

\section{Decomposition Architecture Details}
\label{sec:decomp_arch}

The DecompositionPheromoneModel replaces DPPN's learned soft groupers with fixed spatial segmentation. Key components:

\begin{itemize}[leftmargin=*]
\item \textbf{Regional decomposition:} The sequence of length $N$ is divided into $R = 8$ contiguous regions of size $N/R$. Each region's representation is the average of its token embeddings.

\item \textbf{Sub-claim encoder:} A two-layer MLP ($d \to d/2 \to d$) projects regional embeddings into the pheromone coordinate space.

\item \textbf{Pattern matching:} The average of encoded regional embeddings is compared to $K$ cluster centroids via cosine similarity with temperature-scaled softmax ($T = 0.1$).

\item \textbf{Pheromone field:} $\btau \in \R^{K \times S}$ where $K$ is the number of pattern clusters and $S$ is the number of evaluation strategies. Entry $\tau_{cs}$ records how effective strategy $s$ has been for inputs matching cluster $c$.

\item \textbf{Priority-weighted attention:} Each token's contribution to attention is modulated by its region's priority score, which is derived from the pheromone-advised strategy selection.
\end{itemize}

Test configuration: $d = 64$, $R = 8$ regions, $K = 32$ clusters, $S = 8$ strategies.

\section{Full Per-Seed Results}
\label{app:per_seed}

\begin{table}[h]
\centering
\caption{DPPN transfer results (AULC) by seed and condition.}
\small
\begin{tabular}{lccccc}
\toprule
Seed & Condition & A2 & A3 & B1 & C1 \\
\midrule
\multirow{3}{*}{42}
  & Cold & .683 & .689 & .693 & .676 \\
  & Distilled & .677 & .690 & .695 & .675 \\
  & Rank-red. & .678 & .691 & .695 & .678 \\
\midrule
\multirow{3}{*}{137}
  & Cold & .696 & .697 & .719 & .665 \\
  & Distilled & .706 & .696 & .715 & .661 \\
  & Rank-red. & .707 & .693 & .715 & .666 \\
\midrule
\multirow{3}{*}{256}
  & Cold & .718 & .718 & .733 & .704 \\
  & Distilled & .718 & .717 & .724 & .699 \\
  & Rank-red. & .714 & .715 & .732 & .705 \\
\bottomrule
\end{tabular}
\end{table}

\begin{table}[h]
\centering
\caption{Baseline results (AULC) by seed (cold condition only).}
\small
\begin{tabular}{llcccc}
\toprule
Model & Seed & A2 & A3 & B1 & C1 \\
\midrule
\multirow{3}{*}{Transformer}
  & 42  & .671 & .661 & .669 & .653 \\
  & 137 & .688 & .686 & .689 & .648 \\
  & 256 & .700 & .691 & .711 & .692 \\
\midrule
\multirow{3}{*}{Random Sparse}
  & 42  & .658 & .657 & .671 & .642 \\
  & 137 & .680 & .675 & .703 & .638 \\
  & 256 & .696 & .687 & .684 & .669 \\
\bottomrule
\end{tabular}
\end{table}

\end{document}